\pdfoutput=1
\documentclass[10pt]{article}
\usepackage[final]{EMNLP2023}
\usepackage{times}
\usepackage{latexsym}
\usepackage[T1]{fontenc}
\usepackage[utf8]{inputenc}
\usepackage{microtype}
\usepackage{inconsolata}
\usepackage{graphicx,inputenc,color,algpseudocode,algorithm,comment}
\usepackage{amsmath}

\title{DISGO: Automatic End-to-End Evaluation for Scene Text OCR}
\author{Mei-Yuh Hwang,  Yangyang Shi, \\
Ankit Ramchandani, Guan Pang, Praveen Krishnan, \\
 Lucas Kabela, Frank Seide, Samyak Datta, Jun Liu}
\date{May 2023}

\begin{document}
\maketitle

\begin{abstract}
This paper discusses the challenges of optical character recognition (OCR) on natural scenes, which is harder than OCR on documents due to the wild content and various image backgrounds. We propose to uniformly use word error rates (WER)  as a new measurement for evaluating scene-text OCR, both end-to-end (e2e) performance and individual system component performances. Particularly for the e2e metric, we name it DISGO WER as it considers Deletion, Insertion, Substitution, and Grouping/Ordering errors.
Finally we propose to utilize the concept of super blocks to automatically compute BLEU scores for e2e OCR machine translation.
The small SCUT public test set is used to demonstrate WER performance by a modularized OCR system.

\end{abstract}

\section{Introduction}
OCR on documents is a well studied topic for decades and many useful applications are prevailing such as
receipt scanning and check scanning 
Scene text OCR (ST-OCR) on the other hand, is just getting attraction in recent years, partly due to 
the rise in augmented reality.
Compared to document OCR, scene-text OCR is much harder due to its wild content, with various
image backgrounds. Often text is
scattered in the image and sometimes curved, slanted, or even occluded by other objects. 
If the image is taken by smart glasses, the wearer may be far away
from the object he/she is interested in, and thus the captured text may appear very small. 

Various actions can be executed based on OCR results, such as machine translation (MT), making a call, setting a reminder, etc. Not only do we want to extract text out of the images, we would like OCR to analyze the layout so that it is easier
for the downstream tasks. For example, in Fig \ref{fig:caution}, OCR needs to put at least
"NIÑOS JUGANDO" (children playing) together as one semantic block in that particular reading order.

To evaluate OCR, the existing measurement includes precision, recall, F1, and Panoptic Quality (PQ) \cite{panoptic2019, singh2021textocr, karatzas2013icdar, karatzas2015icdar}. 
While they are all great measurement for detection tasks, it's not as easy to apply them to all of our application interests.
For example, for layout analysis, we need to evaluate both grouping and ordering for the downstream applications.
Hiertext \cite{Long2022} uses PQ for evaluating grouping exclusively, but not ordering. \cite{CTB2022} uses three different metrics to capture different angles
of the e2e performance: local accuracy (similar to the GO error in Section \ref{sec:GO} based on leadership), local continuity (similar to ngram precisions in BLEU), and global accuracy
(measuring exact block accuracy). We follow their contextual text block (CTB) definition for our text layout, where their "integral text units" are
individual words in our system.

While we appreciate the three metrics proposed in \cite{CTB2022}, we advocate the use of word error rates (WER) as a uniform concept across different segments of the system, 
including individual components and e2e measurement with text layout.  Through the location map in Section \ref{sec:location}, one can visualize clearly the error patterns in both detection and recognition. For example, high deletion and insertion errors may imply deficiency in the text detection model, while high
substitution errors are likely due to weak word recognition models.
Finally a separate grouping/ordering WER measures the contextual block accuracy. 
Adding together they evaluate the system e2e, where we name the metric {\bf DISGO} WER, as it
includes all four types of errors: Deletion, Insertion, Substitution, and Grouping/Ordering errors. 
WER is intuitive and easy to understand. Basically it measures how many word errors a system makes, over the number of ground-truth words. It has been widely adopted in
the speech recognition community for decades, and now we've found a way to connect it to images. By analyzing WER distribution in deletions, insertions, substitutions, and
grouping/ordering errors, it helps finding the weakest link in the system 
for a focused and quick improvement.

The innovations include (1) to put each ground-truth word and predicted word onto the image map, (2) in the definition of grouping and ordering errors, and (3) in the generalization of superblock-based sacreBLEU for automatic evaluation from OCR to machine
translation.

We first introduce a baseline ST-OCR system that we put together quickly to demonstrate the feasiblity of  on-device OCR. Section 3 is the focus of this paper, to define WER
for component analysis and e2e measurement, including automatic evaluation on MT output. Section 4 describes two efforts to improve the word recognition component. Experiments in Section 5 show WER results on a small public data set.
 We conclude our satisfaction of our proposed metrics and will be using it to monitor our progress in ST-OCR.


\section{Baseline OCR System}
\label{sec.system}
While there are many advanced OCR technologies that one can apply \cite{liu2021abcnet, liao2020mask, ronen2022glass, huang2022swintextspotter, trOCR}, we aim at carrying out all AI computation on-device for better privacy, connection and latency. To build a prototype quickly, our first system is modularized to the following three major components:
word detection, word recognition, grouping and ordering. Meanwhile we are moving toward more state-of-the-art technologies.

\subsection{Word detection}
We first run word bounding box detection, possibly on down-sampled pixel resolution for latency reasons. 
The bounding box detection model is a Faster-RCNN model with an efficient convolutional backbone optimized for on-device performance \cite{ren2015faster, fbnetv3}. 
The model has 1.68M parameters; it uses post-training quantization and is deployed at 8-bit for all weights and activation functions.  We further use Non-Maximum-Suppression (NMS) to prune overlapped boxes, and a confidence threshold $\alpha$ of 0.6 to prune away low-confidence boxes. Each word box is a rectangular shape and is identified by its center, width, height and a rotation angle $(c_x, c_y, w, h, \theta)$.

\subsection{Word recognition}
\begin{figure}[h!]
    \centering
    \includegraphics[width=7cm]{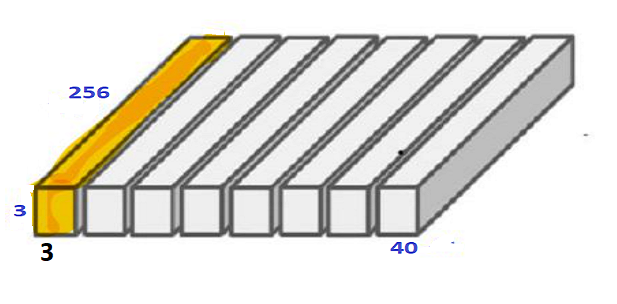}
\begin{center}
\begin{tabular}{ |l|l| } \hline
input image & $320 \times 48 \times 3$  \\ \hline
 final features & $40 \times 3 \times 256$ \\ \hline
 logit & $40 \times 1 \times 111$  \\ \hline
\end{tabular}
\end{center}
    \caption{Final features of word recognition models, via CNN, before logit.}
    \label{fig:word_reco}
\end{figure}

After word bounding box detection, we run word recognition per box independently.  Each word box is warped to horizontal orientation first, and then scaled to a fixed height with its width scaled proportionally.
We use efficient convolutional networks optimized for on-device performance to build the the recognition model \cite{fbnetv2}. In the configuration we ran in Section \ref{sec:exp}, we made each word box at $320 \times 48$ (width $\times$ height) pixels,
and after convolutions, the final feature map is $40 \times 3 \times 256$
 as illustrated in Fig \ref{fig:word_reco}. Finally these features are passed to $3 \times 3 \times 256$ kernel with stride 1, with padding only on the width axis, to 111 channels. Therefore the output is $40 \times 111$, or 40 time-stamps (left to right) of 111-dim logit vectors, which after softmax provide the posterior probabilities of the 111 output units, each of which
represents one character to be recognized.

These characters include English and Spanish letters, punctuations and symbols. Our goal is to have one single recognition model for multiple languages. The model is trained with CTC \cite{Graves2006} loss, and therefore one of the output characters is the CTC blank symbol. Through CTC, the output word is a variable length of at most 40 characters. During inferencing, greedy output is chosen --- the character with the max posterior is picked as the output per time stamp, independent of output from other time stamps. 
Similar to word detection, we have a minimum word recognition confidence pruning, $\beta$, at 0.8.
During training, ground-truth (GT) word boxes are given, and curriculum learning is applied to gradually include longer and longer words across epochs. The recognition model has 1.1M parameters and are quantization-aware trained (QAT), and deployed models are 8-bit compressed. 

\subsection{Grouping and ordering}
In many downstream applications, providing the bag of words recognized is often not enough. For example if we want to pass OCR results to a machine translation system, MT works best given enough context.  Grouping and ordering for scene-text images is non-trivial. We currently have a heuristic-based grouping algorithm to group neighboring word boxes, and then further deciding the reading order among the words in the same group. Each such group is called a block. A block is composed of one or more physical lines. Words within a line have a fixed reading order and lines within a block have a specific reading order as well. This is similar to the definition in the SCUT data set  \cite{CTB2022}.

\section{DISGO: Word Error Rates as e2e Metrics}
WER has been widely adopted by the speech recognition community for many decades. WER is the ratio of the minimum edit distance from the machine output to the ground truth, over the number of GT words. WER can be greater than 1.

We will demonstrate how we can relate OCR metrics with WER. The advantages of using WER for ST-OCR is (1) simple and
easy to understand intuitively: a single number vs. precision/recall/F1 scores, (2) same WER concept across different OCR system components.  High deletion errors implies low recall and high insertion errors implies low precision. By examining Deletion, Insertion, Substitution, and Grouping/Ordering errors (hence the term DISGO for e2e metric), one can adjust each system component based on different application needs.  In addition to an innovative approach of defining WER for grouping and ordering, we will also show how we enable e2e measurement from image input to machine translation output. 

In this paper, we will assume that for each image, we are given $G$ ground-truth words, which are recognized into $P$ predicted words by some OCR system. Furthermore, regardless of how an OCR system is implemented, we assume every GT word is human-annotated with a word box and the OCR system also outputs a word box for each word recognized. This is important when the downstream application needs to know where the recognized word is from, for example, to label it on the image at the detected location. The following proposed metric definition is independent of how an ST-OCR system is implemented, so long as every word has a bounding box.

\subsection{Location map and DIS errors}
\label{sec:location}
The first notation we want to introduce is called a location map. Treat your image as a map, and put all GT words on the map first, based on their bounding box coordinates. Let's call these word locations $1, 2, ... |G|$. This way, each box has a unique ID even if the same word occurs multiple times on the same image, which actually happens often. In this paper, sometimes we use "location" and "word" interchangeably. Sometimes we use "location" specifically to emphasize the unique location on the image map. Readers should be able to tell from the context.

Next for each predicted box, we want to find which GT box it corresponds to, based on intersection-over-union (IoU) with the GT boxes.
We use linear\_sum\_assignment to find the global optimal 1:1 bipartite assignment between P and G:
\begin{center}
{\scriptsize
\begin{verbatim}
from scipy.optimize import linear_sum_assignment
def location_map(g_ocations, G, P, epsilon=1e-5):
    """ Given g_locations[g]=g for g in 1..|G| 
        (g_locations[0] is dummy)
       Return
       (1) p_locations[p] for each p in P, where p=1, 2, .... |P|
       (2) error_codes[x] for all words x in Union(G, P). 
       (3) If error_codes[x] is D, it means G[x] is deleted. 
           Negate g_locations[x]
       (4) If error_codes[x] is I, it means P[x] is inserted. 
           Negate p_locations[x]
       Notice because location 0 is dummy,
           |g_locations| = |G|+1
           |p_locations| = |P|+1
           max(|G|,|P|)+1 <= |error_codes| <= |G|+|P|+1
    """
    Compute IoU(g, p) for each g in G and p in P.
    # IoU.shape == (|G|,|P|), indexed from 0
    idx1, idx2 = 
      scipy.optimize.linear_sum_assignment(IoU, maximize=True)
    assert len(idx1) == min(|G|,|P|) == len(idx2)
    error_codes = [None] * (|G|+1)
    p_locations = [None] * (|P|+1)
    for i, j in zip(idx1, idx2):
        if IoU[i, j] > epsilon:
            g = i+1
            p = j+1
            loc = g_locations[g]
            p_locations[p] = loc
            if G[g] == P[p]: # If the spellings are identical,
              error_codes[loc] = C
            else:
              error_codes[loc] = S
    for g in range(1,|G|+1): 
        if error_code[g] is None:
            error_codes[g] = D
            g_locations[g] = -g_locations[g]
    for p in range(1,|P|+1):
        if p_locations[p] is None:
           next_loc = len(error_codes) # starting from |G|+1
           p_locations[p] = -next_loc
           error_codes.append(I)
    return error_codes
\end{verbatim}
}\vspace*{0.5em}
Algorithm 1: Assigning each predicted word\\a unique location ID.\vspace*{0.5em}
\end{center}

The main idea behind our WER definition is to assign an error code to each location (both GT and prediction locations). Furthermore we will be treating each
location equally, in the sense that a location will be counted as one error at most, even if it is both a substitution error and a grouping/ordering error.

After location\_map in Algorithm 1 is finished, we have an error code assigned to each location. A word in $G$ is either correctly recognized (C) or a substitution error (S) if there is a predicted box overlaps with this GT box. If not, it is a deletion (D) error.

Similarly if a box in $P$ maps to a GT word location, it is either C or S. If it doesn't map to any box in $G$, it is an insertion error (I) and will be assigned to location ID $|G|+1, |G|+2, ...$. Deletion errors often happen to small words, un-focused images, bad lighting, or a weird camera angle. Insertion errors happen when the word detection model proposes too many boxes, either hallucinating word boxes that do not exist at all in GT, or multiple boxes for the same GT box survive all pruning. In the latter case, only the predicted box with the maximum IoU gets assigned to the GT location; the rest, though might also have some overlap with the same GT box, will be considered as insertion errors.

Notice if a location is an insertion or deletion, we eventually negate the location ID in Algorithm 1, as a reminder that it is either not in the GT, or not in the prediction. Only positive locations are in both G and P.

\subsection{Grouping and ordering (GO) errors}
\label{sec:GO}
 A predicted word is considered correct only if (1) the word location matches a GT location, (2) the word spelling matches the GT spelling at the matched location, and (3) the grouping and ordering of the word matches the grouping/ordering of the GT word. We have explained how we calculate errors for the first two conditions. The remaining work now is to decide if a location is in the correct group and in
 the correct reading order.

\begin{center}
\begin{figure}[h]
    \centering
    \includegraphics[width=4.5cm]{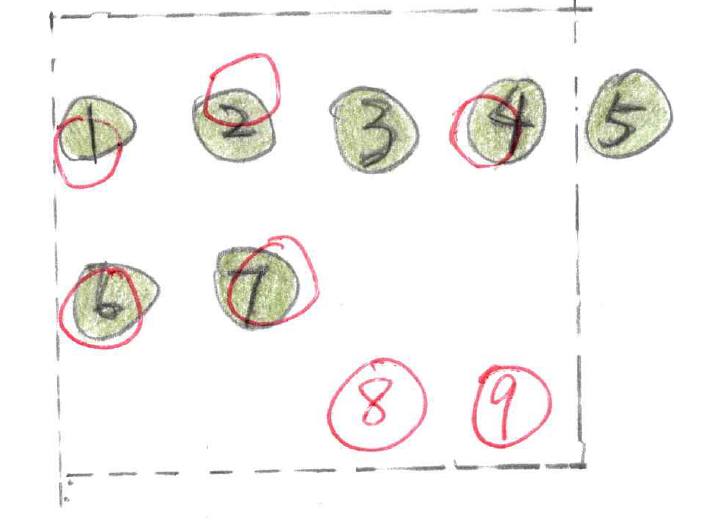}
    \caption{A hypothetical example image, where there are 7 GT words in two blocks (1 2 -3 4 -5) and (6 7) in green. And there are 7 predicted words in red circles, into three blocks (1 2 4 7) (6) (-8 -9).}
    \label{fig:example}
\end{figure}
\end{center}
\vspace{-0.5cm}

To evaluate grouping and ordering errors (abbreviated as GO errors), we need to compare GT block definition vs. machine block definition, and the reading order. We propose the following algorithm to compute GO errors:  
\begin{enumerate}
    \item Represent each block with a location tuple, following the reading order defined by the block. See Fig \ref{fig:example} for examples.
    \item Remove negative locations from the tuple. This is because negative locations in G do not occur in P and vice versa. Hence there is no GO error definition for those negative locations. Besides we already label them as D or I errors, and we don't want to count errors twice for the same location. 
    \item Now the "filtered" GT blocks and the  "filtered" predicted blocks cover the same set of positive locations, whose error\_codes are either C or S.
     {\bf A positive location is considered GO correct if it is following the same leader in the filtered GT block as in the filtered predicted block}.  Notice we don't separate grouping errors vs. ordering errors. They are combined as GO errors. 
\end{enumerate}

The following table computes the leader of each positive location in Fig \ref{fig:example}, where the leader of the first word in a block is defined to be location 0 (indicating the beginning of a block).
\vspace{0.2cm}

    {\small
\begin{tabular}{ |c|c|c| } \hline
 & GT & pred \\ \hline
filtered blocks & (1 2 4) (6 7) & (1 2 4 7) (6) \\ \hline \hline
loc & leader(GT) & leader(pred) \\ \hline
1 & 0 & 0 \\ \hline
2 & 1 & 1 \\ \hline
4 & 2 & 2 \\ \hline
6 & 0 & 0 \\ \hline
7 & 6 & 4 \\ \hline
\end{tabular}
}

\vspace*{0.5cm}
Based on the definition, only location 7 is a GO error. Since we don't want to count errors more than once for any location, error\_codes[7] will be changed to
GO only if it is originally labeled as C by the location map.

\subsubsection{Find bestGT from multiple block definitions}
\label{sec:bestGT}
\begin{center}
\begin{figure}[h!]
    \centering
    \includegraphics[width=3cm]{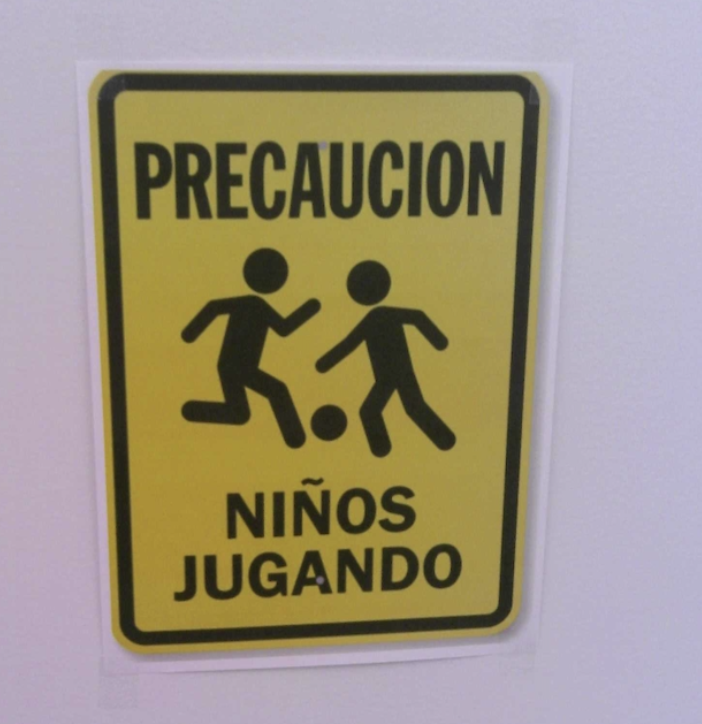}
    \caption{Block definition can be ambiguous at times, as shown in this Spanish poster.}
    \label{fig:caution}
\end{figure}
\end{center}

One minor nuisance about human block definition is that it can be ambiguous at times. For example, Fig \ref{fig:caution} can be two blocks ({\scriptsize "PRECAUCION"} and {\scriptsize "NIÑOS JUGANDO"}) or one block ({\scriptsize "PRECAUCION NIÑOS JUGANDO"}). We should not penalize machines when either block definition is output. Hence we allow multiple GT block annotations per image, similar to MT where a source sentence may have multiple translations.

Assume we are given $K$ human block definitions for an image by $K$ annotators. We can simply compute GO errors based on each of the $K$ block definitions, and pick the one that gives us the minimum GO errors. 

Better yet we should accept combinations of all allowable blocks. As an extreme hypothetical example in Table \ref{tab:block}, there are two block annotators for an image of five words. Annotator A decomposes the image to three blocks, while B two blocks. They also disagree on the ordering of the last block. We see for locations 1, 2, 3, there can be two different block definitions: (1) (2 3) or (1 2 3). For locations 4 and 5, there are also two different block definitions: (4 5) or (5 4). Therefore an OCR system should not be penalized as long as it outputs any of the 4 block definitions for the entire image. Let's use $K' \ge K$ to denote the combinatory number of allowable block definitions. We need an algorithm to find these $K'$ allowable block definitions given $K$ block annotators.

\begin{table}[h!]
\centering{\small
    \begin{tabular}{|l|l|l|l|} \hline
    Annotator A & (1) (2 3) (4 5) & & \\ \hline
    Annotator B & (1 2 3) (5 4) & & \\ \hline \hline
    {\color{blue} EQ ID} & {\color{blue} locations} & {\color{blue} superblock } & {\color{blue} $E_i$}\\ \hline
    000 & 1,2,3 & A=(1) (2 3) & 2   \\ 
      &       & B=(1 2 3) & \\ \hline
    001 & 4,5 & A=(4 5) & 2 \\
     & & B=(5 4) & \\ \hline
    \end{tabular}
    \caption{An extreme hypothetical block disagreement among two annotators, resulting in $K'=2 \times 2=4$ acceptable block definitions.}
    \label{tab:block}}
\end{table}

The algorithm is as simple as finding equivalent classes. An equivalent (EQ) class consists of those locations where there exists an annotator who defines them to be in the same block. Initially all locations of a block from the first annotator form an EQ class. Next we we add one annotator a time to expand the EQ classes until we exhaust all annotators, as Algorithm 2.  Notice all block annotators cover exactly the same set of G words. At this point, we have not put predicted words on the location map yet. Therefore all location IDs are positive: 1, 2, ... $|G|$.

\begin{center}
{\scriptsize
\begin{verbatim}
For each annotator, represent the block annotation as a dict of 
   location id to block tuple, such as 
   loc2b = {1: (1,2,3), 2:(1,2,3), 3:(1,2,3), 4:(5,4), 5:(5,4)}
   
classes = {}
loc2EQ = {}
class_id = 0
for each block of annotator 1:
   classes[class_id] = set(block)
   for loc in classes[class_id]:
      loc2EQ[loc] = class_id
   class_id += 1
   
for annotator k in [2, 3, .. K]:
   Represent annotator k by loc2block dict.
   input_classes = list(classes.keys())
   done = set()
   for each class i in input_classes:
       if i in done:
           continue
       set1 = classes[i]
       merge_us = [i]
       while True:
                set2 = set(loc2 for loc1 in set1 for loc2 in loc2block[loc1])
           if set1 == set2:
              break
           for loc in set2:
               if loc not in set1:
                  class_j = loc2EQ[loc]
                  merge_us.append(class_j)
                  set1 = set1.union(classes[class_j])
            if set1 == set2:
                break
                
        if len(merge_us) > 1:
            classes[class_id] = set1
            for loc in classes[class_id]:
                 loc2EQ[loc] = class_id
            class_id += 1
            for d in merge_us:
                del classes[d]
        done = done.union(merge_us)
\end{verbatim}}
Algorithm 2: Finding equivalence classes
\end{center}

Once the EQ classes are identified, we check how each annotator defines blocks for all the locations inside each EQ class (see "superblock" column in Table \ref{tab:block}). Notice that an EQ class may imply more than one block for some annotators, hence the term "superblock".

Assume there are $E_i$ different block definitions among all $K$ annotators for EQ class $i$, then
$ K' = \prod_i E_i $. 
Referring to Table \ref{tab:block}, $K' = 2 \times 2  = 4$. 

Next we put predicted words on the location map. Since now we may have negative locations, we update these $K'$
block definitions by negating those deleted locations.
Then we go through all of $K'$ definitions to select the one with the minimum GO errors as our {\bf bestGT} block definition. 
To re-iterate: we treat each location equally. So if a positive location is a substitution error (error code is S) already, we don't count it
as an additional GO error even if its grouping/ordering is wrong. In other words, GO is labeled only for those locations
where the error codes were C. As GO does not affect deletion, insertion, and substitution errors, (D, I, S) error codes in Algorithm 1 remain the same after grouping/ordering evaluation is applied.

We have now completed OCR e2e metric (from image input to block text output) as
{\color{blue} WER(e2e) = DISGO = (D+I+S+GO)/$|G|$}. 

If one would like to understand how frequent grouping/ordering is wrong 
among all positive locations, let's introduce
GS as the number of locations whose grouping/ordering is wrong and whose error codes are S.
Then the frequency of making grouping/ordering mistakes is {\color{blue} WER(GO) = (GO+GS) / (C+S)}

\subsection{Evaluating OCR components separately}
In practice we often develop each system component in parallel. Enabling the evaluation of each system component alone is critical to allow
independent and simultaneous development of multiple components.
Here we show it's easy to use WER as component metrics.

\begin{itemize}

\item To evaluate the word detection component by itself, NMS is still applied, but no $\alpha$ confidence pruning.  Based on the location map without string comparison (no S errors since no word recognition is run), {\color{blue} WER(detection) = (D+I)/$|G|$}, with IoU $\epsilon=0.5$ in Algorithm 1.  High deletions correspond to low recall, while high insertions correspond to low precision. Such a direct and intuitive WER can replace precision/recall metrics. 

\item To evaluate the word recognition component alone, we usually assume GT bounding boxes are given. Given that we run recognition using GT word boxes without any $\beta$ confidence pruning, we can define {\color{blue}WER (recognition) = (S+D)/$|G|$ }. 
Deletions are possible in our case because CTC may output the blank symbol for all timestamps, which results in an empty word. 

\item To evaluate grouping/ordering errors alone, we are given the GT word boxes (and the GT word strings if the grouping/ordering
algorithm needs them), hence all error codes are C in Algorithm 1. 
We follow Section \ref{sec:GO} to compute GO errors and obtain {\color{blue}WER(grouping\&ordering) = GO/$|G|$}.

\end{itemize}

\subsection{Evaluating end-to-end machine translation}
If the application scenario is to send OCR blocks to machine translation, the app will be calling MT API per {\color{blue} OCR-defined block}. In order to run MT system evaluation automatically, the evaluation data contain human translation for each {\color{blue} GT block}. A problem thus incurs: How do we find the GT translation for each OCR-defined block so that we can evaluate MT automatically?

Inspired by  EQ classes among all block annotators, now we just need to find
the EQ classes between bestGT in Section \ref{sec:bestGT} and the predicted block definition. The major difference is that bestGT and predicted blocks may cover different sets of locations: Some GT locations might be deleted and OCR might have inserted new locations.
 These deleted and inserted locations have negative location IDs. It's a minor yet tricky modification of Algorithm 2 to accommodate negative locations, where we treat bestGT as the first annotator and prediction the 2nd annotator. 

Furthermore after the 2nd annotator (the machine) is added to expand the initial EQ classes,
 there might be leftover blocks in prediction that contain purely negative locations (all insertions such as EQ 002 in Table \ref{tab:ordering}). We have to create a separate EQ class per inserted block. 

In general, each EQ class contains a mapping of "many blocks" to "many blocks". Table \ref{tab:ordering} shows the super blocks computed from aligning bestGT and prediction in Fig \ref{fig:example}. It also shows the human translation (HT) of each GT block, where $a_1$ has two  translations and $a2$ has 3 translations. The machine generates "i love yes" as the translation of block $b_1$, and "fine" for $b_2$, "the the the" for $b_3$.

\begin{table*}[h!]
\centering
{\scriptsize
    \begin{tabular}{|l|l|l|} \hline
     & EQ 001 & EQ 002 \\ \hline\hline
     bestGT & $a_1 = (1, 2, {\color{red}-3}, 4, {\color{red}-5}),  a_2=(6, 7) $ & \\ 
            & HT($a_1$)=["i love you dearly", "i am very fond of you"] & \\
            & HT($a_2$)= ["fine", "all right", "yes"] & \\ \hline
     prediction & $b_1=(1, 2, 4, 7),  b_2=(6)$ & $b_3={\color{red}(-8, -9)}$\\ 
                & MT($b_1$) = "i love yes" & MT($b_3$) = "the the the" \\
                & MT($b_2)$ = "fine" & \\ \hline
    \end{tabular}}
    \caption{Superblocks from aligning (bestGT, prediction) in Fig \ref{fig:example}.}
    \label{tab:ordering}
\end{table*}

Review BLEU definition from \cite{papineni2002},
\begin{eqnarray*}
    \log \mbox{BLEU} & = & \min(1-\frac{r}{c}, 0) + \sum_{i=1}^N w_i \log p_i \\
    p_i & = & \frac{\mbox{\# i-gram hits with GT translations}}{\mbox{\# i-grams in MT output}}
\end{eqnarray*}
where $p_i$ is the i-gram precision, $r$ is the total {\em best-match-{\color{red} r}eference-length} of the test set, and $c$ is the total length of machine translation of the test set (c represents "{\color{red}c}orpus"). The first item is the brevity penalty (bp) as precision measurement has penalized long translations already. Brevity penalty penalizes short translations (smaller $c$ making the first item a negative number). BLEU is the bigger the better. $N=4$ and $w_i=1/N$ is the most common BLEU researchers have reported. 

Conceptually we will be treating each superblock as one sentence in traditional MT, and compute the N-gram hits per superblock. {\bf The main difference is that we don't count ngrams across block boundaries within a superblock}, respecting the fact that
there is no sequential relationship across blocks. It's easy to modify sacreBLEU \cite{sacreBLEU} to compute our superblock-level 
ngram precisions in order to compute BLEU.

\subsubsection{Multiple ground-truth translations per block}
To deal with multiple GT translations per GT block, assume there are $h$ GT blocks in a superblock and each block $i$ has $F_i$ GT translations. we will pretend this superblock has  $\prod_{i=1}^h F_i$ translations while computing the set of GT ngrams.
For $[a_1, a_2]$ superblock in Table \ref{tab:ordering},  we will use $ 2 \times 3 = 6$ GT translations to compute
N-gram hits. Again it's an easy modification of sacreBLEU.

Table \ref{tab:bleu} computes BLEU for the two images in Fig \ref{fig:caution} (assuming
GT has only one block definition) and Table \ref{tab:ordering}. Here we follow case insensitive sacreBLEU with
 with use\_effective\_order=True and smooth="exp".
Since $c > r$, bp=1. BLEU = $(\frac{7}{10}\frac{2}{6}\frac{1}{2 \times 3})^{(1/3)} = 33.88$. 

\begin{table}[h!]
\centering
{\scriptsize
\begin{tabular}{ |l|c|c|c|c| } \hline
Superblock & ngram hits & all ngrams & c & r \\ \hline
GT=["caution", "children playing"] & & & & \\
Pred=["caution children playing"] & [3 1 0 0] & [3 2 1 0] & 3 & 3 \\
\hline
GT=[a1 a2], Pred=[b1 b2] & [4 1 0 0] & [4 2 1 0] & 4 & 5 \\ 
\hline
GT=[], Pred=[b3] & [0 0 0 0] & [3 2 1 0] & 3 & 0 \\
\hline \hline
TOTAL & [7 2 0 0] & [10 6 3 0] & 10 & 8 \\ \hline
\end{tabular}}
\caption{BLEU computation for the two images in Fig 4 and Table 2.}
\label{tab:bleu}
\end{table}

Table \ref{tab:ordering} is an 
example where a GT superblock is empty. Our algorithm also takes care of the other condition when the GT superblock is not empty while the prediction superblock is.

\section{Incorporating Linguistic Knowledge}
\def\LM{\mathrm{LM}}
\def\CTC{\mathrm{CTC}}
The CTC-based word recognizer decodes each characters independently. This can lead to jarring errors such as transcribing ``ground'' as ``gr0und''. To repair such deficiency, we supplement the image 
recognition system with linguistic knowledge via a language model $P_\LM(W)$ that models the probability of the character sequence $W$.

The recognizer output is the sequence $W$ with the highest posterior probability $P(W | X)$ given the input image $X$. $P(W | X)$ is obtained via Bayes' rule:
{\scriptsize
\begin{eqnarray}
P(W | X) & = & P(X | W) \cdot P_\LM(W) / P(X) \nonumber \\
& = & \frac{{\color{blue}P_\CTC(W | X)}P_\CTC(X)}{\color{blue}P_\CTC(W)} \cdot {\color{blue}P_\LM(W)} /P(X) \nonumber \\
& \approx & {\color{blue} \frac{P_\CTC(W | X)}{P_\CTC(W)} P_\LM(W)} \quad \label{eq:sll}
\end{eqnarray}
}
$P_\CTC(W | X)$ is the CTC image model output.
The terms $P_\CTC(X)$ and ${P(X)}$ do not affect the decision and thus are ignored in Eq \ref{eq:sll}.
The prior $P_\CTC(W)$ is computed as the product of character unigram probabilities.
Beam search decoding is used to find the sequence of characters with the maximum posterior in Eq \ref{eq:sll}.
While applying the prior term in the denominator for large-vocabulary speech recognition is usually unimportant, we find
it brings us some improvement consistently.
 For the LM itself, we experimented with a character-based N-gram and an LSTM LM. Future work will extend this to language modeling across words within a block.

A second approach to incorporating linguistic information is through Transformer-based sequence-to-sequence modeling \cite{vaswani2017attention}, where the decoder learns linguistic knowledge while predicting the next character.
For rapid experimentation, we project the logits in Fig \ref{fig:word_reco} from a dimension of 111 to 256
as the input to our transformer. The encoder has four self-attention layers of hidden dimension of 256, with eight attention heads (abbreviated as L4A8H256), and feed-forward dimension of 1024 (abbreviated as FF1024). The decoder is L2A8H256 and FF1024. The decoder
vocabulary keeps the same 111 characters (though it never outputs the blank symbol), in addition to three
new symbols: <bos>, <eos>, and PADDING.
Starting from <bos> as the initial input, the decoder keeps generating the next character until <eos> is output.  The Transformer model has 5.7M parameters, in addition to the 1.1M parameters in the baseline.

\begin{figure}
    \centering
    \includegraphics[width=7.5cm]{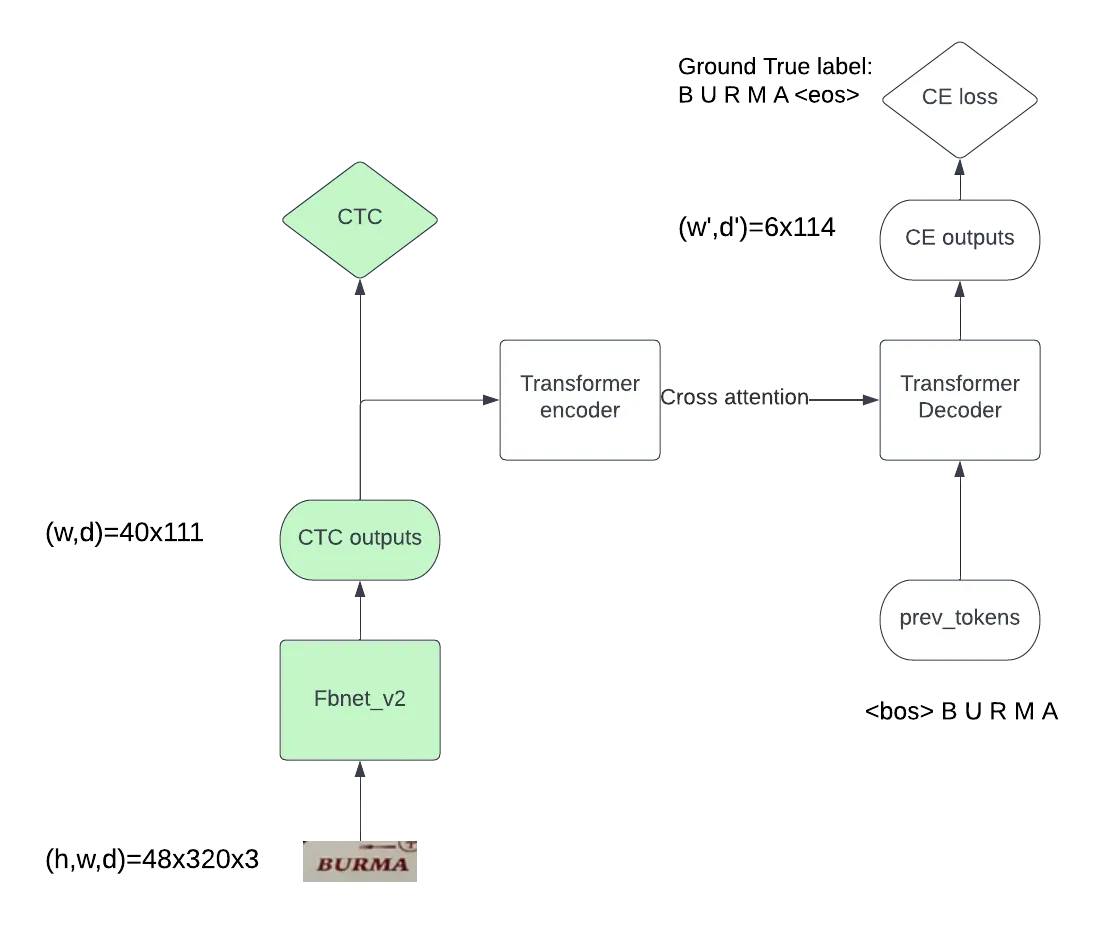}
    \caption{Transformer architecture for word recognition}
    \label{fig:transformer}
\end{figure}

As Fig \ref{fig:transformer} shows, training is based on multi-task learning (MTL) between the original CTC loss and the Transformer cross-entropy (CE) loss. Both losses back-propagate to update the CNN image feature network. In the case of training on the word "BUMRA", the decoder will generate 6 posterior vectors of dimension 114 for CE training.
After the model is converged, one may also extract the CTC subnet as an improved baseline.

\section{Experiments \& Discussion}
\label{sec:exp}
 Our OCR training and validation data set are extracted from TextOCR \cite{singh2021textocr} and Intel OCR \cite{krylov2021open} corpora, plus
some additional internal data we collected.
It includes both English and Spanish words, for a total of 3M training words and 118k words for validation.
For training character-level LMs, we use the text from all the image training data, 
plus some additional text from various images, for a total of 5.5M words.

Since the focus of this paper is to present WER as ST-OCR end-to-end metric, we use a small public test set to  illustrate the merits of the metrics, the SCUT-CTW1500 test set \cite{scut2017,CTB2022}, denoted as SCUT-test. This challenging data set 
consists of 497 images with mostly
curved text in the wild, with block annotation. 
Excluding <unk> words and words containing characters outside our 110 alphabets, there are 4938 words. Detection/recognition models are 8-bit quantized.

\subsection{E2E baseline results}
\label{sec:bl}
Since SCUT-test images have low resolution (around $500 \times 375$ range), we don't down sample the input image resolution for word detection.
The e2e error distributions are shown in Table \ref{tab:baseline}, where C, D, I, S is the number of correctly recognized words,
deletion errors, insertion errors, and substitution errors calculated in Algorithm 1, before GO errors are applied.
"GO" is the number of correct words that got turned into GO errors by Section \ref{sec:GO}.
Similarly "GS" is for those substitution errors turning into GO errors. 

\vspace{0.3cm}
\begin{table}
\centering
   {  \scriptsize
    \begin{tabular}{|l|c|c|c|c|r|r|r|} \hline
   $\beta$ &   C   & D & I & S & GO & GS & D+I+S\\ \hline\hline
    0.8 &     2390 & 1921 & 94 & 627 & 533 & 157 & 2642\\ \hline
    0 &     2453 & 1462 & 184 & 1023 & 573 & 274 & 2669 \\ \hline \hline
    diff & +63 & -459 & +90 & +396 & & & +27 \\ \hline
    \end{tabular}}
{\scriptsize
    \begin{verbatim}
    beta = 0.8
    G = C+S+D = 4938
    P = C+S+I = 3111
    WER(DIS) = (D+I+S)/G = 53.5%
    WER(GO)  = (GO+GS)/(C+S) = 22.9%
    WER(e2e) = (D+I+S+GO)/G = 64.30%
       C': 1857/4938=37.61%  (C'=C-GO)
        D: 1921/4938=38.90%
        I: 94/4938=1.90%
        S: 627/4938=12.70%
       GO: 533/4938=10.79%
\end{verbatim}

\caption{Error distribution of SCUT test set by the baseline OCR system}
    \label{tab:baseline}
}
\end{table}

We see that there is a large amount of deletion errors, even without any pruning by the recognition model ($\beta$=0, D=1462/4938=29.6\%). 
Confidence $\beta$=0.8 pruned 549 boxes. 459 of them mapped to some GT words, with 63 correctly recognized
and 396 incorrectly recognized. The remaining 90 boxes are forged by the detection model and did not map to any GT word.

Now that
we have the baseline system, and the metrics are in place, we will be digging into each system component for improvement,
and perhaps change the entire flow to a completely e2e transformer. We also notice our heuristic-based grouping and ordering
does not work well on curved text. Machine learning based grouping and ordering, such as \cite{CTB2022,Long2022,Cruz2017,Noroozi2016} is the solution.

\subsection{LM results}
WER=(S+D)/$|G|$ with language modeling are shown in Table \ref{tab:results}, for word recognition given ground-truth bounding boxes.
No confidence pruning is used, and hyper-parameters are pre-tuned on a separate internal validation set. 

\begin{table}\centering
{\scriptsize
    \begin{tabular}{|c|l|c|c|r|c|} \hline
System & Model & Quant. & S & D & WER \\ \hline\hline
A & Baseline & Y & 1410 & 34 & 29.2\% \\ \hline
B & A + 5-gram LM & Y+N & 1212 & 0 & 24.5\% \\ \hline
C & A + LSTM LM & Y+N & 1165 & 0 & {\bf 23.4\%} \\ \hline
D & A + Transformer & Y+N &1169 &0 & 23.7\% \\ \hline
E & CTC subnet 1 & N & 1234 & 5 & 25.1\% \\ \hline
F & CTC subnet 2 & Y & 1291 & 16 & 26.5\% \\ \hline
    \end{tabular}}
\caption{Adding language models on word recognition given ground-truth word bounding boxes.}
\label{tab:results}
\end{table}

Row 1 in Table \ref{tab:results} shows the baseline without LM. Notice the 34 deletion errors were caused by the CTC blank symbol being chosen for every character output. In Row 2, we added a 5-gram char-level LM. Among the pre-tuned hyper parameters for the decoder, there was a reward for character insertions. Due to the reward, we did not observe deletion errors with the 5-gram LM.
The LSTM in System C consisted of 3 LSTM layers with a hidden width of 512, followed by a linear layer and softmax. Similarly no deletion error was observed.
System D is the Transformer output, with beam size 5, while greedy output (beam=1) yielded 9 more errors. Transformers do not generate the blank symbol. Though there is a slim possibility that the decoder may generate <eos> as the first output character, we did not observe it in this experiment, hence no deletion error as well. Overall it achieved about the same performance as LSTM, with slightly fewer parameters.

System E extracted the transformer's CTC subnet directly and used as is, reducing the WER from 29.24\% to 25.09\%. To understand
how much we lost from quantization, we took System E as the initial model and re-trained the recognition model using QAT training for System F.
We see quantization causing relatively 5.5\%=$|E-F|/E$ of degradation. Yet System F was still 9.5\% relatively better than the baseline, with identical neural network architecture and quantization. This shows MTL training helps improving the baseline network \footnote{We observed similar behavior regarding quantization among various baseline systems, and MTL impact on the CTC subnets, using other big test sets.}.

Finally Table \ref{tab:params} lists the additional parameters in System B, C, D, along with the test-set perplexity when an explicit LM was used. LSTM  had a lower perplexity than the 5-gram and
performed slightly better as well, though at the price of three times of parameters. Notice none of these additional parameters was quantized.

\begin{table}\centering
 { \scriptsize
    \begin{tabular}{|l|c|c|} \hline
Model & \#params & PPL (SCUT-test) \\ \hline\hline
5-gram & 2.1M & 8.84 \\ \hline
LSTM & 6.4M & 8.18 \\ \hline
Transformer & 5.7M & - \\ \hline
    \end{tabular}}
    \caption{Number of additional model parameters and test-set perplexity.}
    \label{tab:params}
\end{table}

\section{Conclusion and Future Work}
This paper presents a unified metric for evaluating scene-text OCR individual components and end-to-end performance, including evaluating all the way through machine translation.
The innovations include (1) to put each ground-truth word and predicted word onto the image map, (2) to define grouping and ordering errors, and (3) to generalize sacreBLEU at the superblock level. The metric from image input
to OCR block output is named DISGO, as it includes errors from Deletion, Insertion, Substitution and Grouping/Ordering errors.
This WER metric is easy to understand, and by examining the distribution of the four types of errors,
one can further focus on the critical component for a quick improvement. We hope this metric definition will be widely accepted
in the scene-text OCR field. 

We are continuing on improving our system across multiple directions, including data, model architecture, ML-based
grouping and ordering.
We will continue to evaluate our progress with the WER metrics proposed in this paper.

\section{Limitations}

While the proposed unified metric, DISGO, and the accompanying innovations are valuable contributions to evaluating scene-text OCR performance, there are several limitations to consider. First of all, we have to yet validate the metrics' effectiveness by conducting and comparing with human evaluation on a wide range of datasets and OCR systems, best across different languages, fonts, image qualities, and text complexities. 

Secondly the metric relies on the word bounding box information for word-level alignment between ground-truth words and predicted words.
While most OCR systems do output word bounding boxes (though may not always be rectangular), some models such as transformer-based use "soft" boundaries based on
attentions. It's a bit hard to define  word bounding boxes precisely. Besides, it requires special software for the human annotator 
to draw the ground-truth bounding boxes. It's not as easy to just upload an image and type in the list of words in the image, then the evaluation could be conducted.
To facilitate the latter case when ground-truth bounding boxes are not available, we did internally create the location map by comparing word strings and define IoU as
IoU($g_1$, $p_1$) = 1 - min\_edit\_distance($g_1$, $p_1$)/max($|g_1|$, $|p_1|$). When two IoUs, IoU($g_1$, $p_1$) and IoU($g_1$, $p_2$), are the same, we then compute IoU(B($g_1$),  B($p_1$)) and IoU(B($g_1$),  B($p_2$)) as the secondary
comparison, where B($x$) is the block string where the word $x$ belongs to. This enables us to run quick evaluation on random pictures
we took in the office without
the heavy annotation of bounding boxes, as an approximation on the word alignment. It worked well on images with sparse text. But it still often fails to align to the right ground-truth words on images with dense text with many short words and repeated words.


\section{Ethics Statement}

 An unified metric for evaluating scene-text OCR individual components and end-to-end performance is proposed in this paper. It provides a comprehensive and valuable tool to evaluate scene-text OCR individual components and end-to-end performance. We take steps to minimize the potential negative impacts of our research by experimenting with our hypothesis on a public dataset that contains layout information and thus can facilitate downstream machine translation studies. However, if the method will be applied to sensitive data, such as medical patients, privacy-preserving policies should be taken into account. For example, only non-private and user agreed data can be revealed in 
 the ground-truth, and hence the evaluation should be limited to these non-sensitive data, rather than the entire image.

\bibliographystyle{IEEEbib}
\bibliography{library}

\begin{thebibliography}{10}
\providecommand{\url}[1]{#1}
\csname url@samestyle\endcsname
\providecommand{\newblock}{\relax}
\providecommand{\bibinfo}[2]{#2}
\providecommand{\BIBentrySTDinterwordspacing}{\spaceskip=0pt\relax}
\providecommand{\BIBentryALTinterwordstretchfactor}{4}
\providecommand{\BIBentryALTinterwordspacing}{\spaceskip=\fontdimen2\font plus
\BIBentryALTinterwordstretchfactor\fontdimen3\font minus
  \fontdimen4\font\relax}
\providecommand{\BIBforeignlanguage}[2]{{%
\expandafter\ifx\csname l@#1\endcsname\relax
\typeout{** WARNING: IEEEtran.bst: No hyphenation pattern has been}%
\typeout{** loaded for the language `#1'. Using the pattern for}%
\typeout{** the default language instead.}%
\else
\language=\csname l@#1\endcsname
\fi
#2}}
\providecommand{\BIBdecl}{\relax}
\BIBdecl

\bibitem{panoptic2019}
\BIBentryALTinterwordspacing
A.~Kirillov, K.~He, R.~Girshick, C.~Rother, and P.~Dollár, ``Panoptic
  segmentation,'' in \emph{IEEE/CVF Conference on Computer Vision and Pattern
  Recognition}, 2019, p. 9404–9413. [Online]. Available:
  \url{https://arxiv.org/abs/1801.00868}
\BIBentrySTDinterwordspacing

\bibitem{singh2021textocr}
A.~Singh, G.~Pang, M.~Toh, J.~Huang, W.~Galuba, and T.~Hassner, ``Textocr:
  Towards large-scale end-to-end reasoning for arbitrary-shaped scene text,''
  in \emph{Proceedings of the IEEE/CVF conference on computer vision and
  pattern recognition}, 2021, pp. 8802--8812.

\bibitem{karatzas2013icdar}
D.~Karatzas, F.~Shafait, S.~Uchida, M.~Iwamura, L.~G. i~Bigorda, S.~R. Mestre,
  J.~Mas, D.~F. Mota, J.~A. Almazan, and L.~P. De~Las~Heras, ``Icdar 2013
  robust reading competition,'' in \emph{2013 12th international conference on
  document analysis and recognition}.\hskip 1em plus 0.5em minus 0.4em\relax
  IEEE, 2013, pp. 1484--1493.

\bibitem{karatzas2015icdar}
D.~Karatzas, L.~Gomez-Bigorda, A.~Nicolaou, S.~Ghosh, A.~Bagdanov, M.~Iwamura,
  J.~Matas, L.~Neumann, V.~R. Chandrasekhar, S.~Lu \emph{et~al.}, ``Icdar 2015
  competition on robust reading,'' in \emph{2015 13th international conference
  on document analysis and recognition (ICDAR)}.\hskip 1em plus 0.5em minus
  0.4em\relax IEEE, 2015, pp. 1156--1160.

\bibitem{Long2022}
S.~Long, S.~Qin, D.~Panteleev, A.~Bissacco, Y.~Fujii, and M.~Raptis, ``Towards
  end-to-end unified scene text detection and layout analysis,''
  \emph{https://arxiv.org/abs/2203.15143}, 2022.

\bibitem{CTB2022}
\BIBentryALTinterwordspacing
C.~Xue, J.~Huang, S.~Lu, C.~Wang, and S.~Bai, ``Contextual text block detection
  towards scene text understanding,'' \emph{European Conference on Computer
  Vision}, 2022. [Online]. Available:
  \url{https://arxiv.org/pdf/2207.12955.pdf}
\BIBentrySTDinterwordspacing

\bibitem{liu2021abcnet}
Y.~Liu, C.~Shen, L.~Jin, T.~He, P.~Chen, C.~Liu, and H.~Chen, ``Abcnet v2:
  Adaptive bezier-curve network for real-time end-to-end text spotting,''
  \emph{IEEE Transactions on Pattern Analysis and Machine Intelligence},
  vol.~44, no.~11, pp. 8048--8064, 2021.

\bibitem{liao2020mask}
M.~Liao, G.~Pang, J.~Huang, T.~Hassner, and X.~Bai, ``Mask textspotter v3:
  Segmentation proposal network for robust scene text spotting,'' in
  \emph{Computer Vision--ECCV 2020: 16th European Conference, Glasgow, UK,
  August 23--28, 2020, Proceedings, Part XI 16}.\hskip 1em plus 0.5em minus
  0.4em\relax Springer, 2020, pp. 706--722.

\bibitem{ronen2022glass}
R.~Ronen, S.~Tsiper, O.~Anschel, I.~Lavi, A.~Markovitz, and R.~Manmatha,
  ``Glass: Global to local attention for scene-text spotting,'' in
  \emph{Computer Vision--ECCV 2022: 17th European Conference, Tel Aviv, Israel,
  October 23--27, 2022, Proceedings, Part XXVIII}.\hskip 1em plus 0.5em minus
  0.4em\relax Springer, 2022, pp. 249--266.

\bibitem{huang2022swintextspotter}
M.~Huang, Y.~Liu, Z.~Peng, C.~Liu, D.~Lin, S.~Zhu, N.~Yuan, K.~Ding, and
  L.~Jin, ``Swintextspotter: Scene text spotting via better synergy between
  text detection and text recognition,'' in \emph{Proceedings of the IEEE/CVF
  Conference on Computer Vision and Pattern Recognition}, 2022, pp. 4593--4603.

\bibitem{trOCR}
\BIBentryALTinterwordspacing
M.~Li, T.~Lv, L.~Cui, Y.~Lu, D.~A.~F. Flor{\^{e}}ncio, C.~Zhang, Z.~Li, and
  F.~Wei, ``Trocr: Transformer-based optical character recognition with
  pre-trained models,'' \emph{CoRR}, vol. abs/2109.10282, 2021. [Online].
  Available: \url{https://arxiv.org/abs/2109.10282}
\BIBentrySTDinterwordspacing

\bibitem{ren2015faster}
S.~Ren, K.~He, R.~Girshick, and J.~Sun, ``Faster r-cnn: Towards real-time
  object detection with region proposal networks,'' \emph{Advances in neural
  information processing systems}, vol.~28, 2015.

\bibitem{fbnetv3}
\BIBentryALTinterwordspacing
X.~Dai, A.~Wan, P.~Zhang, B.~Wu, Z.~He, Z.~Wei, K.~Chen, Y.~Tian, M.~Yu,
  P.~Vajda, and J.~E. Gonzalez, ``Fbnetv3: Joint architecture-recipe search
  using neural acquisition function,'' \emph{CoRR}, vol. abs/2006.02049, 2020.
  [Online]. Available: \url{https://arxiv.org/abs/2006.02049}
\BIBentrySTDinterwordspacing

\bibitem{fbnetv2}
\BIBentryALTinterwordspacing
A.~Wan, X.~Dai, P.~Zhang, Z.~He, Y.~Tian, S.~Xie, B.~Wu, M.~Yu, T.~Xu, K.~Chen,
  P.~Vajda, and J.~E. Gonzalez, ``Fbnetv2: Differentiable neural architecture
  search for spatial and channel dimensions,'' \emph{CoRR}, vol.
  abs/2004.05565, 2020. [Online]. Available:
  \url{https://arxiv.org/abs/2004.05565}
\BIBentrySTDinterwordspacing

\bibitem{Graves2006}
A.~Graves, S.~Fernandez, F.~Gomez, and J.~Schmidhuber, ``Connectionist temporal
  classification : Labelling unsegmented sequence data with recurrent neural
  networks,'' \emph{Proc. ICML}, 2006.

\bibitem{papineni2002}
\BIBentryALTinterwordspacing
K.~Papineni, S.~Roukos, T.~Ward, and W.-J. Zhu, ``Bleu: a method for automatic
  evaluation of machine translation,'' \emph{ACL}, pp. 311--318, 2002.
  [Online]. Available: \url{https://aclanthology.org/P02-1040.pdf}
\BIBentrySTDinterwordspacing

\bibitem{sacreBLEU}
\BIBentryALTinterwordspacing
M.~Post, ``A call for clarity in reporting {BLEU} scores,'' in
  \emph{Proceedings of the Third Conference on Machine Translation: Research
  Papers}.\hskip 1em plus 0.5em minus 0.4em\relax Belgium, Brussels:
  Association for Computational Linguistics, Oct. 2018, pp. 186--191. [Online].
  Available: \url{https://www.aclweb.org/anthology/W18-6319}
\BIBentrySTDinterwordspacing

\bibitem{vaswani2017attention}
A.~Vaswani, N.~Shazeer, N.~Parmar, J.~Uszkoreit, L.~Jones, A.~N. Gomez,
  {\L}.~Kaiser, and I.~Polosukhin, ``Attention is all you need,''
  \emph{Advances in neural information processing systems}, vol.~30, 2017.

\bibitem{krylov2021open}
I.~Krylov, S.~Nosov, and V.~Sovrasov, ``Open images v5 text annotation and yet
  another mask text spotter,'' in \emph{Asian Conference on Machine
  Learning}.\hskip 1em plus 0.5em minus 0.4em\relax PMLR, 2021, pp. 379--389.

\bibitem{scut2017}
L.~Yuliang, J.~Lianwen, Z.~Shuaitao, and Z.~Sheng, ``Detecting curve text in
  the wild,'' \emph{https://arxiv.org/abs/1712.02170}, 2017.

\bibitem{Cruz2017}
R.~S. Cruz, B.~Fernando, A.~Cherian, and S.~Gould, ``Deeppermnet: Visual
  permutation learning,'' \emph{IEEE Conference on Computer Vision and Pattern
  Recognition}, 2017.

\bibitem{Noroozi2016}
M.~Noroozi and P.~Favaro, ``Unsupervised learning of visual representations by
  solving jigsaw puzzles,'' \emph{European Conference on Computer Vision},
  2016.

\end{thebibliography}
\end{document}